%% file: socs19.tex
\pdfoutput=1    
\documentclass[letterpaper]{article}
\usepackage{aaai19}

\usepackage{{algorithmic}}
\usepackage{times}
\usepackage{helvet}
\usepackage{courier}
\usepackage{natbib}
\usepackage{url}
\usepackage{enumitem}
\usepackage{siunitx}
\usepackage{amssymb}
\usepackage{amsmath}
\usepackage[makeroom]{cancel}
\usepackage{mathrsfs}
\usepackage{amsthm}

\usepackage{subfigure}

\usepackage{bm}

\usepackage{graphicx}

\usepackage{multirow}
\usepackage[table,xcdraw]{xcolor}

\usepackage[linesnumbered,vlined,ruled]{algorithm2e}
\makeatletter
\newcommand\notsotiny{\@setfontsize\notsotiny{7}{8}}
\makeatother
\SetKwProg{Fn}{Function}{}{}
\SetAlFnt{\scriptsize}
\SetAlCapFnt{\normalfont\scriptsize}
\SetAlCapNameFnt{\scriptsize}
\SetArgSty{textrm}
\SetInd{0.1em}{0.5em}
\SetCommentSty{emph}
\usepackage{setspace}

\usepackage{tabularx, booktabs}
\usepackage{varioref}
\usepackage[noabbrev,capitalise]{cleveref}

\theoremstyle{definition}

\theoremstyle{definition}

\usepackage{thmtools}

\declaretheoremstyle[%
  spaceabove=0pt,
  spacebelow=0pt,
  headfont=\normalfont\itshape,%
  postheadspace=1em,%
  qed=\qedsymbol%
]{mystyle}

\graphicspath {{figures/}}
\allowdisplaybreaks

\newcommand{\open}{\mbox{OPEN}\xspace}

\newcommand{\focal}{\mbox{FOCAL}\xspace}

\newcommand{\ignore}[1]{}

\begin{document}

\title{Position Paper: From Multi-Agent Pathfinding to Pipe Routing}
\author{
{Gleb Belov},\textsuperscript{\rm 1}
{Liron Cohen},\textsuperscript{\rm 2}
{Maria Garcia de la Banda},\textsuperscript{\rm 1}
{Daniel Harabor},\textsuperscript{\rm 1}
{Sven Koenig}, \textsuperscript{\rm 2}
{Xinrui Wei} \textsuperscript{\rm 1}\\
\textsuperscript{1}Monash University, Australia\\
\textsuperscript{2}University of Southern California\\
\{gleb.belov, daniel.harabor\}@monash.edu,
skoenig@usc.edu}
\date{\today}

\maketitle

\input abstract
\input introduction
\input pipe_routing
\input cbs

\input benchmarks
\input experiments
\input discussion

\input relatedwork
\input conclusion

\small
\bibliographystyle{aaai}
\bibliography{references}

\end{document}

%% file: abstract.tex
\begin{abstract}
The 2D Multi-Agent Path Finding (MAPF) problem aims at finding collision-free
paths for a number of agents, from a set of start locations to a set of goal
positions in a known 2D environment. MAPF has been studied in theoretical
computer science, robotics, and artificial intelligence over several decades,
due to its importance for robot navigation. It is currently experiencing
significant scientific progress due to its relevance in automated warehousing
(such as those operated by Amazon) and in other contemporary application
areas.  In this paper, we demonstrate that many recently developed MAPF
algorithms apply more broadly than currently believed in the MAPF research
community. In particular, we describe the 3D Pipe Routing (PR) problem, which
aims at placing collision-free pipes from given start locations to given goal
locations in a known 3D environment. The MAPF and PR problems are similar:
a solution to a MAPF instance is a set of blocked cells
in x-y-t space, while a solution to the corresponding PR instance is a 
set of blocked cells in x-y-z space. We show how to use this similarity to apply
several recently developed  MAPF
algorithms to the PR problem, and discuss their performance on abstract PR
instances. We also discuss further research necessary to tackle real-world 
pipe-routing instances of interest to industry today. This opens up a new 
direction of industrial relevance for the MAPF research community.
\end{abstract}

%% file: introduction.tex
\section{Introduction}
\label{sec:introduction}
\begin{figure}[t]
    \centering
    \includegraphics[width=0.8\columnwidth]{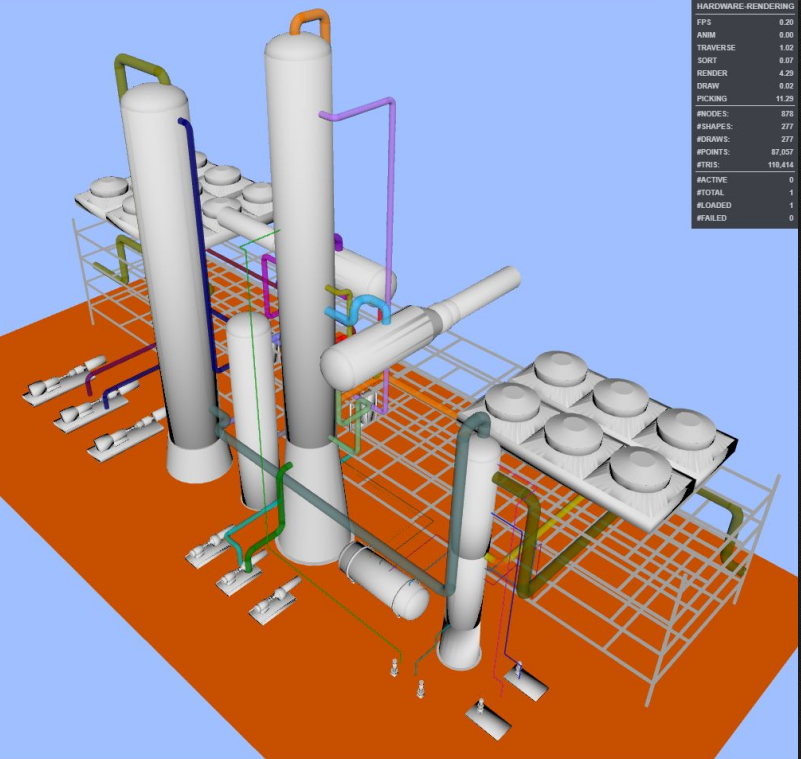}
    \caption{Example layout for the acid gas removal module in a liquefied natural gas plant}
    \label{fig:AGRU}
\end{figure}

The 3D Pipe Routing (PR) problem is a common industrial problem that appears when
designing the layout of industrial plants, such as oil refineries, 
natural gas processing stations, water treatment
facilities, and the type of
power plants used in ships and submarines.
Designing the layout of such a plant requires finding  
3D  location  coordinates  for  every piece of  equipment in the plant 
(equipment allocation problem), and finding a 3D route
for every pipe that connects two pieces of equipment (PR problem). 
The aim is to minimize the total
cost of the plant (which can run into multi-billion dollar budgets), while ensuring safety
and correct functionality. Figure~\ref{fig:AGRU} shows a layout for part of a natural gas plant.

Differences in the quality of the final layout, can have a very significant 
impact on the cost of these plants, including the cost of the pipes and 
associated support structures, which are known to take the largest share: up 
to 80\% of the purchased equipment cost or 20\% of the fixed-capital 
investment~\citep{PetersTimmerhaus1991}.  However, finding high-quality plant 
layouts is remarkably difficult due to the size of these plants and the 
complexity of the associated constraints. As a result,
layouts are still designed manually, taking multiple engineers many months (or
even years) to complete. This process is inefficient, costly and the 
results may vary in quality, since they largely depend on the experience of
the piping and layout engineers.

Current research into automatic plant layout commonly divides it into two phases. The  first  phase  performs equipment allocation, that is, finds 3D positions for all equipment that minimise a total cost and satisfy all equipment constraints, such as min/max distances and maintenance access requirements. In this phase the cost of the pipes is approximated using
rough measures, such as Manhattan distances. The second
phase solves the PR problem, that is, finds 3D routes for all pipes connecting 
the (already allocated) equipment, that minimize the pipe costs (based on their length) and satisfy  all pipe constraints,
such as no two pipe routes collide and they are all appropriately supported.
In such setting, the start and end position of each pipe is given as input 
(referred to as \emph{nozzles}), representing the pipe's connection to its source/target equipment. 

The  PR problem is similar to the  2D  Multi-Agent  Path  Finding  (MAPF)  problem, which searches for
 collision-free  paths  for  several agents  from
given start locations to given goal locations in a known 2D
environment. Thus, a  solution  to  a  MAPF  instance  is  a
set  of  blocked  cells  in  x-y-t  space,  while  a  solution  to  the
corresponding PR instance is the corresponding set of blocked cells
in x-y-z space. In this paper, we show  how  to  use  this  similarity  to  apply
several recently developed optimal and bounded-suboptimal MAPF algorithms to the PR problem. %
We also discuss the performance of some of these algorithms on different sets of PR instances. Finally, we discuss further research necessary to tackle
real-world pipe-routing instances of interest to industry today.
This opens up a new direction of industrial relevance for the
MAPF research community.



%% file: pipe_routing.tex
\begin{figure}[!t]
    \centering
    \includegraphics[width=0.8\columnwidth]{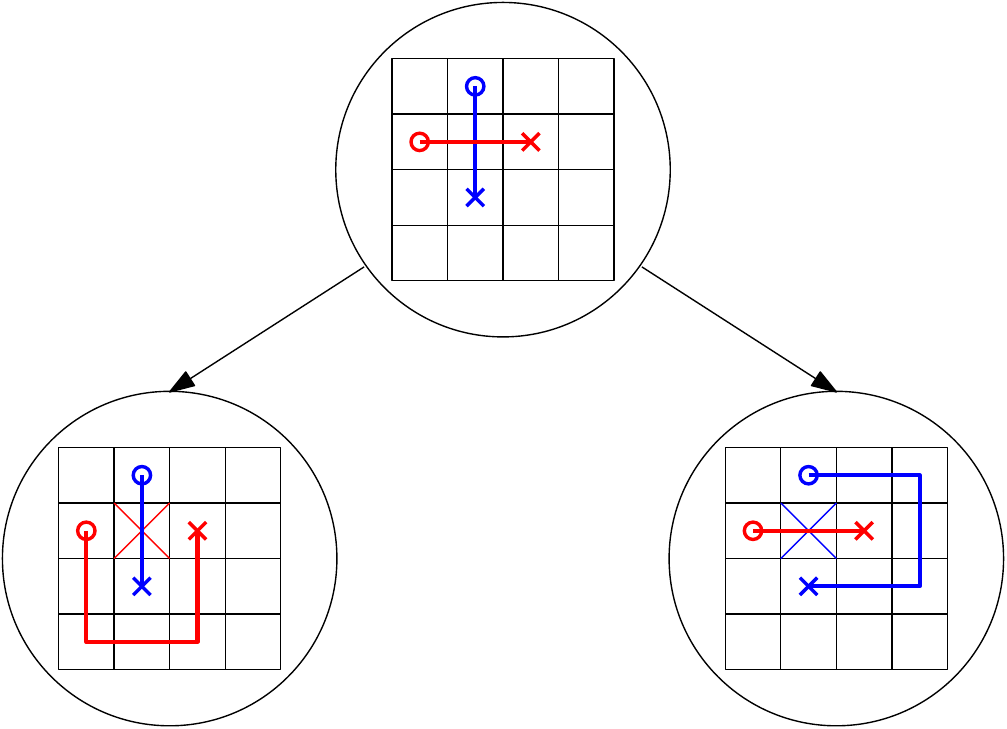}
    \caption{Illustrates Conflict-Based Search for a simple 2D environment with red and blue pipes. `o's and `x's represent start and goal vertices, respectively. The root node has one conflict which is resolved by constraining one of the agents. Both leaves represent feasible solutions of equal cost.}
    \label{fig:HL_Tree}
\end{figure}

\section{The Core Pipe Routing Problem}

This section defines an abstract version of the PR problem, which will be used throughout the rest of the paper. The abstraction uses a discretization of the plant's volume into a unit-cost 6-neighbor 3D grid graph. In this graph, each vertex is associated with a coordinate $(x,y,z) \in Z^3$, and represents either an open or a blocked grid cell at that coordinate. A unit-cost edge between two vertices exists only if both vertices are open. Along with this graph, we are given a set $K$ of pipes. Each pipe $k\in K$ specifies its start and goal vertices, $v^k_S$ and $v^k_G$, respectively. A solution is a set of $K$ routes connecting each pipe's start and goal vertices. A route is simply a path (that is, a sequence of edges) in the abstract graph. A solution is feasible if each vertex (and thus each edge) is used to route at most one pipe. The cost of a solution is the sum of route lengths. Finding a solution with minimum cost is NP-hard (due to the equivalence with 2D MAPF for the objective function of the sum of individual costs, which is known to be NP-hard, \citealt{YL:AAAI:13}).

While the above formulation abstracts away some aspects of the real-world problem (such as pipe stress and support requirements) it maintains the core combinatorial hardness. This is because in a feasible solution pipes may need to accommodate each other and, thus, routed differently than their individual shortest paths.



%% file: cbs.tex
\section{MAPF Algorithms for Core Pipe Routing}
\label{sec:cbs}

\input alg_ecbs

In this paper we solve Core Pipe Routing using two different MAPF algorithms.
%
The first one is \textbf{Conflict-Based Search (CBS)} \citep{SSFS:AIJ:15}, an optimal MAPF solver that performs high-level and low-level searches. Each high-level node contains a set of constraints and, for each agent (we will use the terms agent and pipe interchangeably), a feasible path that respects these constraints. A high-level node is a \emph{goal} node if and only if none of its paths collide. The high-level root node has no constraints. The high-level search of CBS is a \emph{best-first} search that defines the $f$-value of a high-level node as their cost, that is, as the sum of the travel times along the pipe paths it contains. Figure~\ref{fig:HL_Tree} shows an example. 
When CBS expands a high-level node $N$, it checks whether $N$ is a goal node.  If it is, CBS terminates successfully and outputs the paths in $N$ as solution. Otherwise, at least two paths collide. CBS chooses a collision to resolve and generates two high-level child nodes of $N$.
Both child nodes inherit the constraints of $N$ and each has an additional constraint that resolves the chosen collision. During the generation of the high-level node $N$, CBS performs a low-level search for the agent $i$ affected by the newly added constraint. The low-level search for agent $i$ is a (best-first) A* search that ignores all other agents and finds a minimum-cost path from the start vertex of agent $i$ to its goal vertex that respects the constraints of $N$ that involve agent $i$.

The second algorithm is \textbf{ECBS($w$)}~\citep{BSSF:SOCS:14}, a $w$-suboptimal variant of CBS whose high-level and low-level searches are \emph{focal} searches \citep{PJ:IEEE:82}, rather than best-first searches. Like A*, focal searches use an \open list of nodes sorted in increasing order of their $f$-values. 
Unlike A*, focal searches with suboptimality factor $w$ also use a \focal list of all nodes currently in \open, whose $f$-values are no larger than $w$ times $f_{\min}$, the smallest $f$-value in the current \open. The nodes in \focal are sorted in increasing order according to \emph{secondary} heuristic values. While A* expands a node in \open with the smallest $f$-value, a focal search expands a node in \focal with the smallest secondary heuristic value. Secondary heuristic values do not have to be consistent (or even admissible). The high-level and low-level focal searches of ECBS($w$) use measures related to the number of collisions, as secondary heuristic values. Algorithm~\ref{alg_hl} shows a version of ECBS($w$) adapted to the Core Pipe Routing problem by removing the temporal aspect.

%% file: alg_ecbs.tex
\begin{algorithm}[t]
\label{alg_hl}
\footnotesize
\KwIn{Core Pipe Routing problem, $w \geq 1$.}
\KwOut{A $w$-suboptimal solution.}
Init root node with a plan for each agent using focal search.\\
$Push$ the root node to \open and \focal.\\
\While{$\focal \neq \emptyset$}{
  $N \leftarrow Pop(\focal)$.\\
  \DontPrintSemicolon
  \lIf{$N$ is a solution}{
    \KwRet $N$.
  }
  Identify a conflict between agents $j$ and $k$ at cell $c$.\\
  Generate two successor nodes, $N^j$ and $N^k$ for agents $j$ and $k$, each imposing the additional constraint $c$.\\
  Replan using focal search for agents $j$ and $k$ in $N^j$ and $N^k$.\\
  $Push$ $N^j$ and $N^k$ to \open and conditionally to \focal.\\
  $Update$ \focal if necessary.\\
}
\KwRet no solution.
\caption{ECBS (high-level search) adapted to Core Pipe Routing}
\label{alg_hl}
\end{algorithm}

%% file: benchmarks.tex
\section{Experimental Setup}
\label{sec:setup}
For our feasibility study we generate a wide range of Core Pipe
Routing Problem instances, and solve them using variants of some well known MAPF algorithms.
Our experiments attempt to capture one or more of the main challenges
associated with routing pipes in practice: many pipes, many components that
must be avoided, small congested spaces and a requirement that paths be
computed with a high degree of precision.

\paragraph{Environments:}
Our test environments comprise three-dimensional {\it cubes} made up of square {\it cells}.
We generate two different types of cubes and two different sizes for each, all as follows:

\begin{itemize}
\item  {\it Empty environments:}
these are a small cube of size $20 \times 20 \times 20$ and a large cube of size $320 \times 320 \times 320$. 
Each cell in each empty cube is considered traversable. The smaller cube is selected to evaluate performance and 
scalability (in terms of number of pipes) in congested spaces. The larger cube, meanwhile, is selected to evaluate
settings of realistic size, which often contain many millions of cells to explore.

\item {\it Obstacle environments: } these are again two cubes of the same size as the 
empty environments, where we introduce random obstacles.
These environments are selected to evaluate performance and scalability in settings
where there exist many pieces of equipment that must be avoided.
Obstacles take the form of columns, which originate from the floor of the environment 
(i.e. from a cell at height = 0), and rise up 
to a randomly selected height. 
The origins of these columns are also selected at random to comprise 10\% of the cells
in each cube. 
\end{itemize}

\paragraph{Instances: } For each environment we select at random pairs of traversable cells 
(resp. start and target locations) from among those appearing on the perimeter of each cube. 
In other words, we assume pipes begin and end from random positions on the face of each cube. 
A set of such start-target pairs is called an \emph{instance} of the Core Pipe Routing Problem.
The number of pipes per instance varies depending on the environment. For the small environments 
we proceed in increments of 10, up to 200 pipes. For large environments, we proceed in 
increments of 50, up to 650 pipes.

Note that, as with MAPF, two instances having the same number of pipes (equiv. agents) 
may require very different amounts of computation to solve. 
Depending on the positions of the start and target cells one instance may be relatively
straightforward while another can require significant coordination effort.
To mitigate such bias in our results, we choose to generate {\it 50 instances per increment}.

\paragraph{Algorithms:}
We test three suitably modified MAPF algorithms on the set of generated problem instances.
These are:
\begin{itemize}
\item Conflict-based Search (CBS), which returns an optimal solution if one exists.
\item ECBS(1.01), which returns a solution guaranteed to be no more than 1\% larger than optimal.
\item ECBS(1.05), which returns a solution guaranteed to be no more than 5\% larger than optimal.
\end{itemize}

All algorithms are implemented in C++ and have a 100s timeout (i.e. they return
failure if a solution cannot be found after this amount of time).
We repeat each experiment 50 times for different number of
pipes, using different randomly generated start and goal locations.
Our test machine is a 3.5GHz AMD Ryzen 3 2200G desktop computer with 16GB dual channel 
2400MHz RAM.

%% file: experiments.tex
\section{Results}
\label{sec:experiments}
We evaluate each of the three algorithms --- CBS, ECBS(1.01) and ECBS(1.05) --- on each of our four test environments. 
We analyse results as part of two distinct experiments: small environments and large environments.
In each case we focus on three distinct metrics:
\begin{itemize}
\item {\bf Success Rate}, which measures the percentage of problems that were solved for each fixed number of pipes (recall that we generate 50 instances per increment).
We will say that an algorithm {\it usually succeeds} if its success rate for $k$ pipes is $\ge 50\%$.
\item {\bf Runtime}, which measures, in seconds, the time required to solve an instance. Rather than computing average performance (e.g. per $k$ pipes) we focus instead 
on performance across the full set of instances on a given environment. The distribution of results is always presented in sorted order,
from lowest runtime to highest runtime (recall that we use a 100s timeout).
\item {\bf Solution Quality}, which measures the quality of solutions computed by ECBS(1.01) and ECBS(1.05) with respect to the best known bound
(i.e. either the solution cost found by optimal CBS or the minimum $f$-value of any node in the FOCAL list).
\end{itemize}

\subsection{Experiment 1: Small Environments}
\begin{figure*}
\vspace{-1em}
\begin{minipage}{\textwidth}
    \hspace*{-1em}
    \begin{minipage}{.37\textwidth}
    \includegraphics[width=\textwidth]{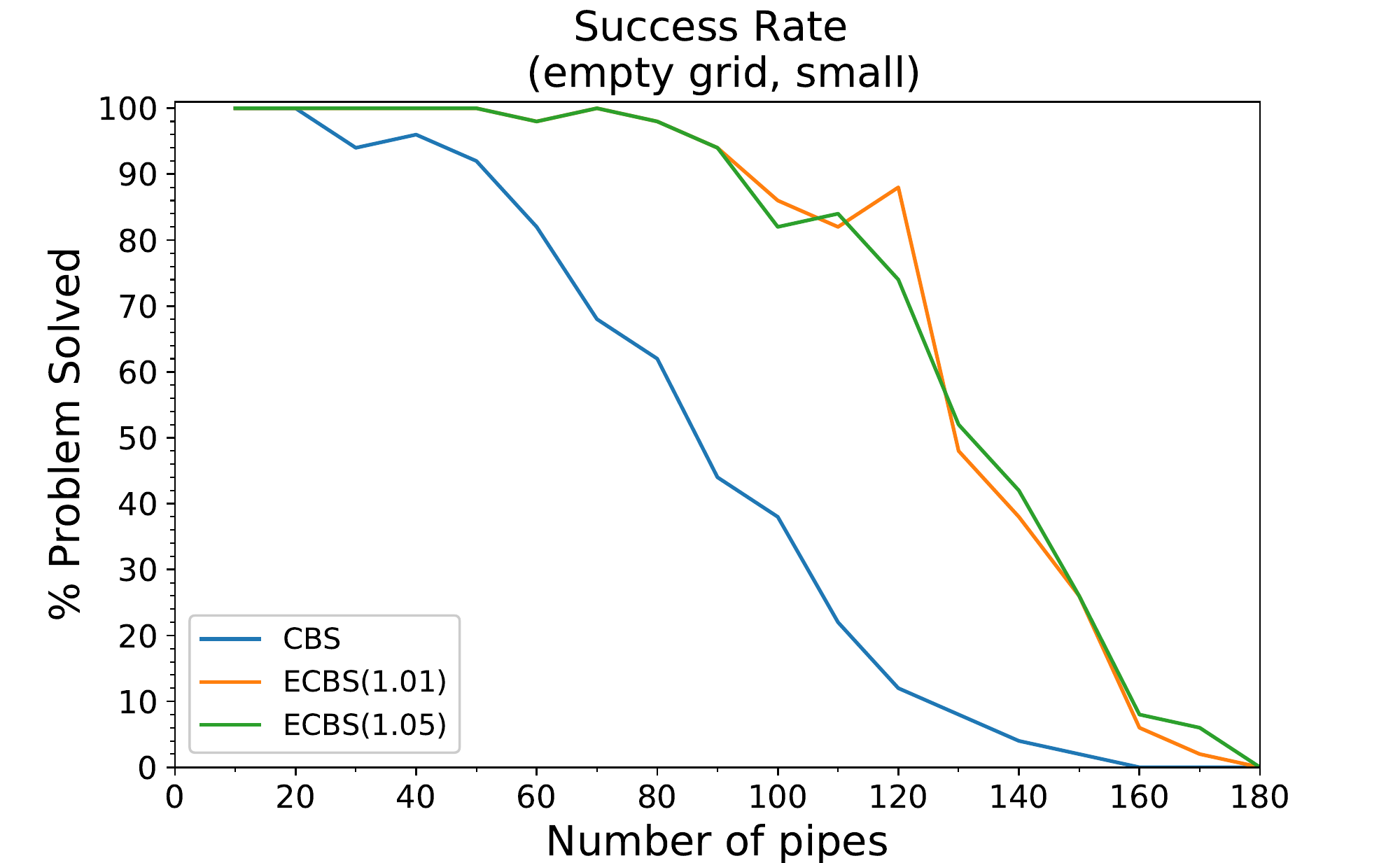}
    \begin{center}
    (a)
    \end{center}
    \end{minipage}
    \hspace*{-1.8em}
    \begin{minipage}{.37\textwidth}
    \includegraphics[width=\textwidth]{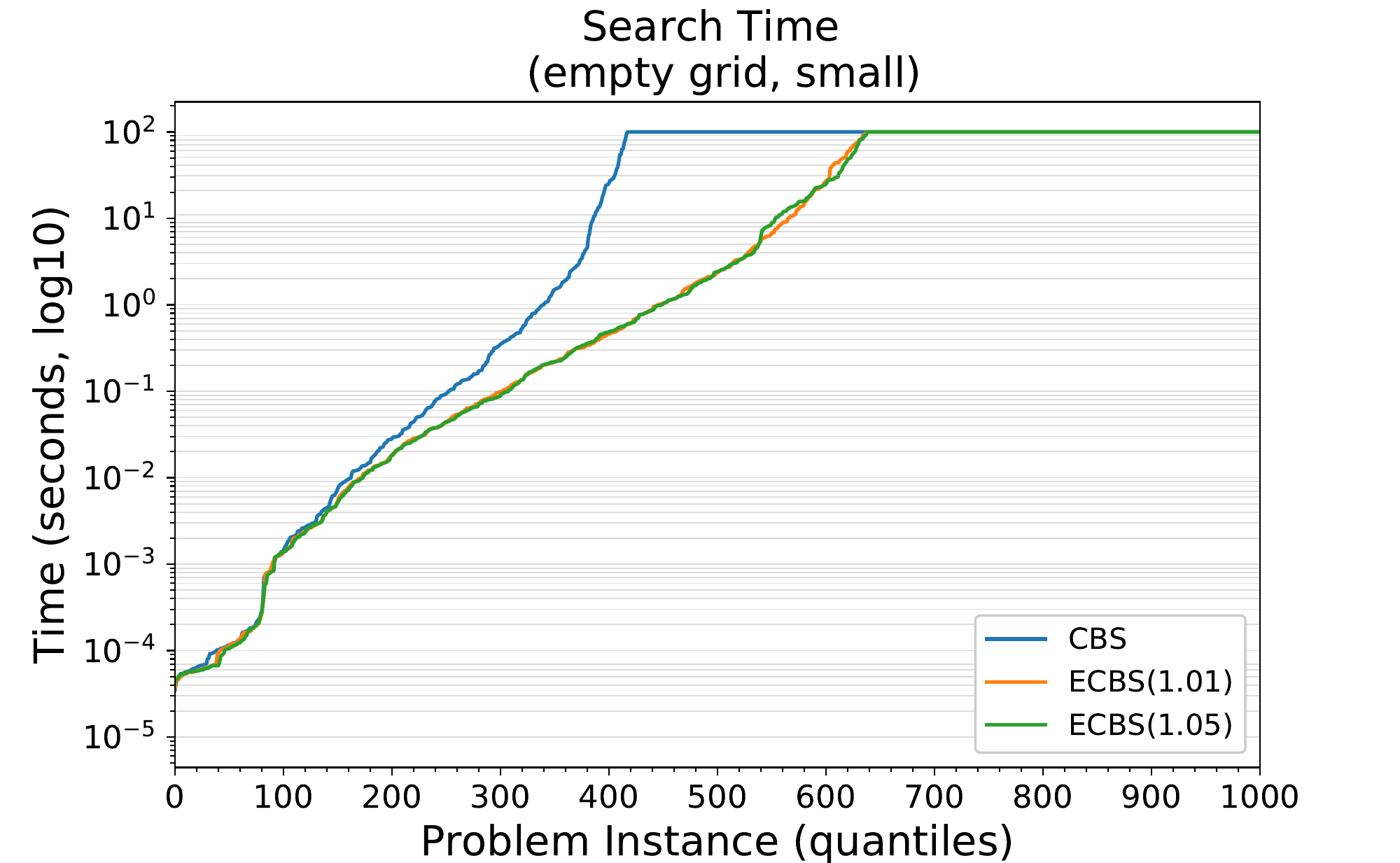}
    \begin{center}
    (b)
    \end{center}
    \end{minipage}
    \hspace*{-1.8em}
    \begin{minipage}{.37\textwidth}
    \includegraphics[width=\textwidth]{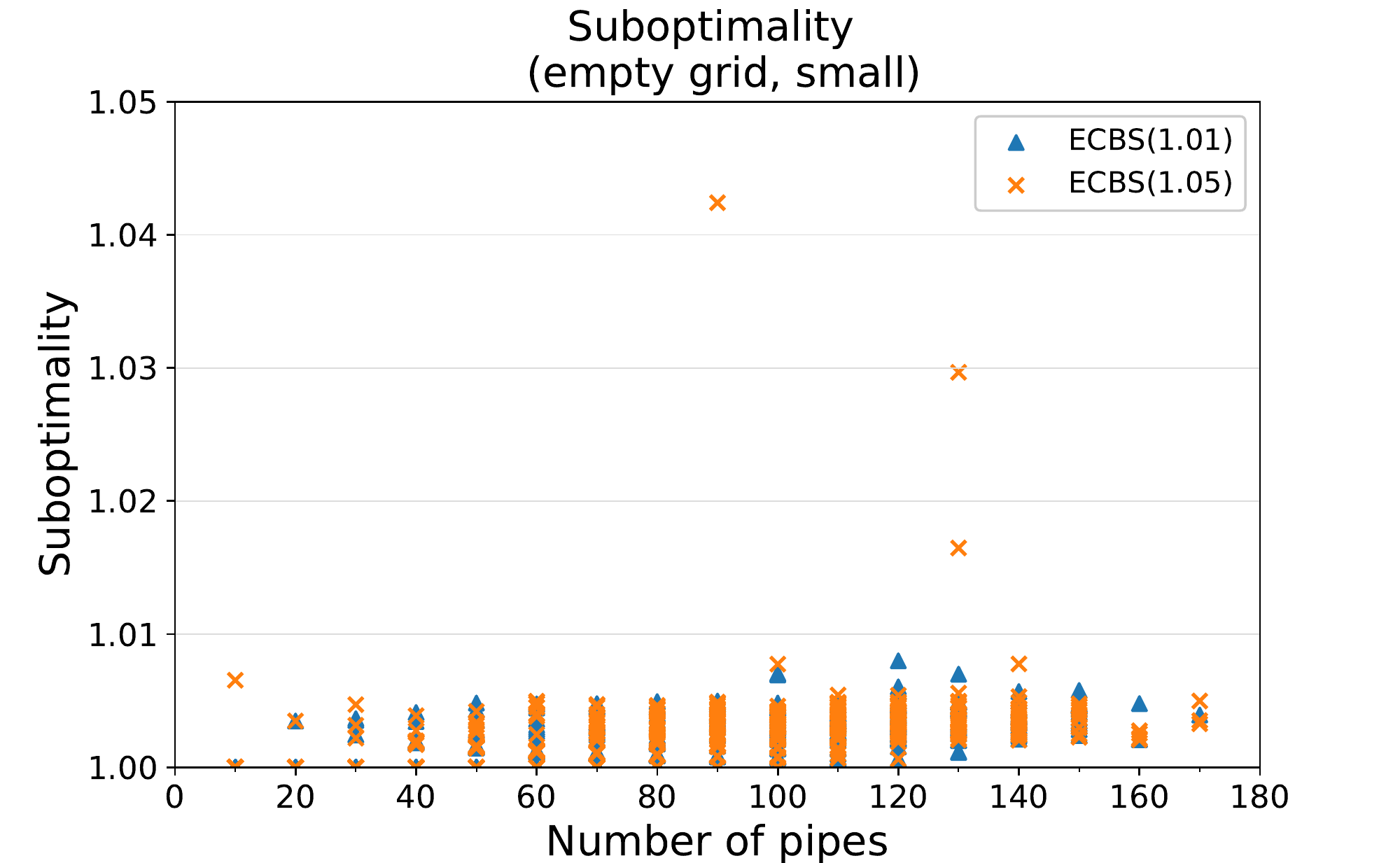}
    \begin{center}
    (c)
    \end{center}
    \end{minipage}
    \hspace*{-1em}
    \begin{minipage}{.37\textwidth}
    \includegraphics[width=\textwidth]{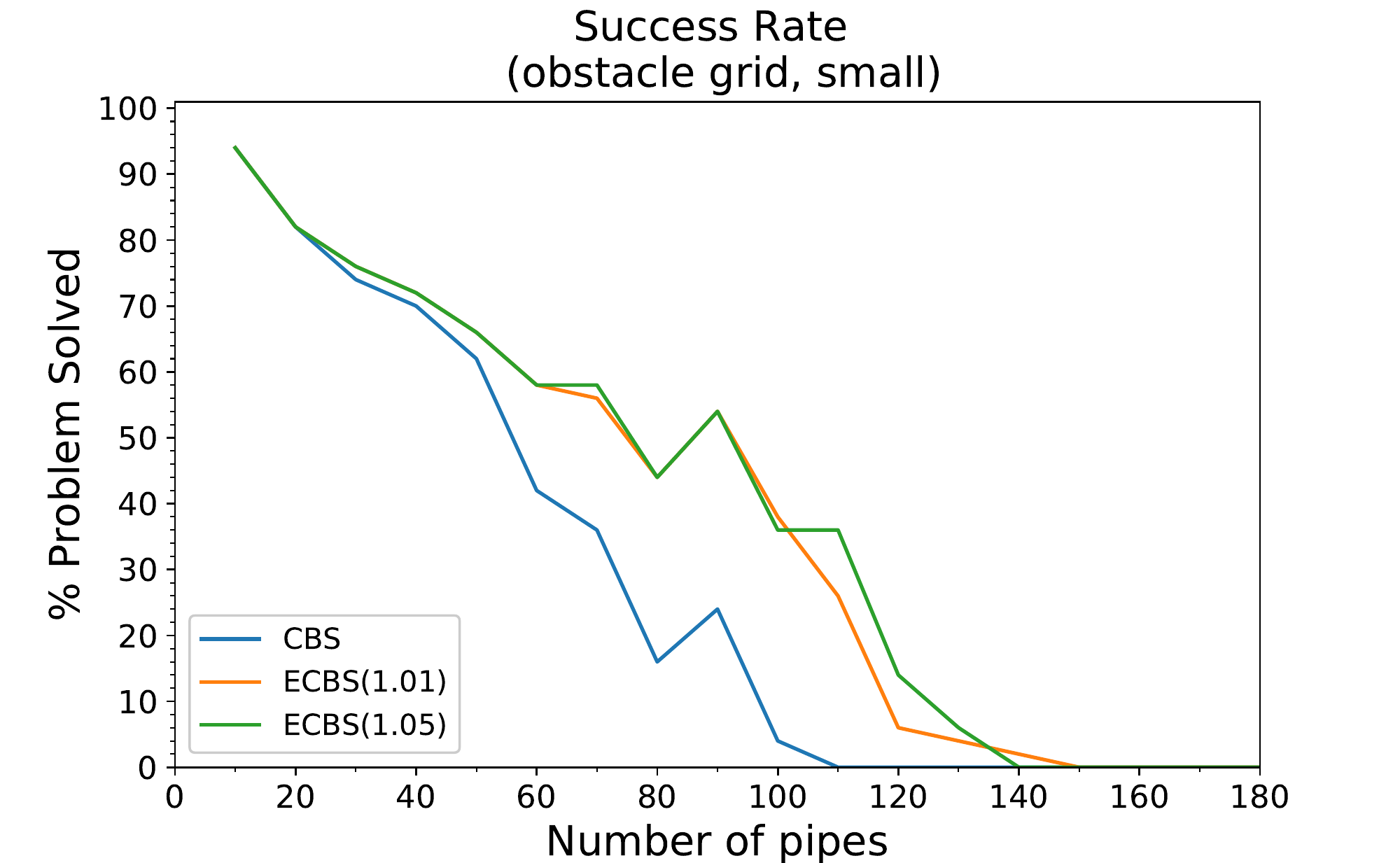}
    \begin{center}
    (d)
    \end{center}
    \end{minipage}
    \hspace*{-1.8em}
    \begin{minipage}{.37\textwidth}
    \includegraphics[width=\textwidth]{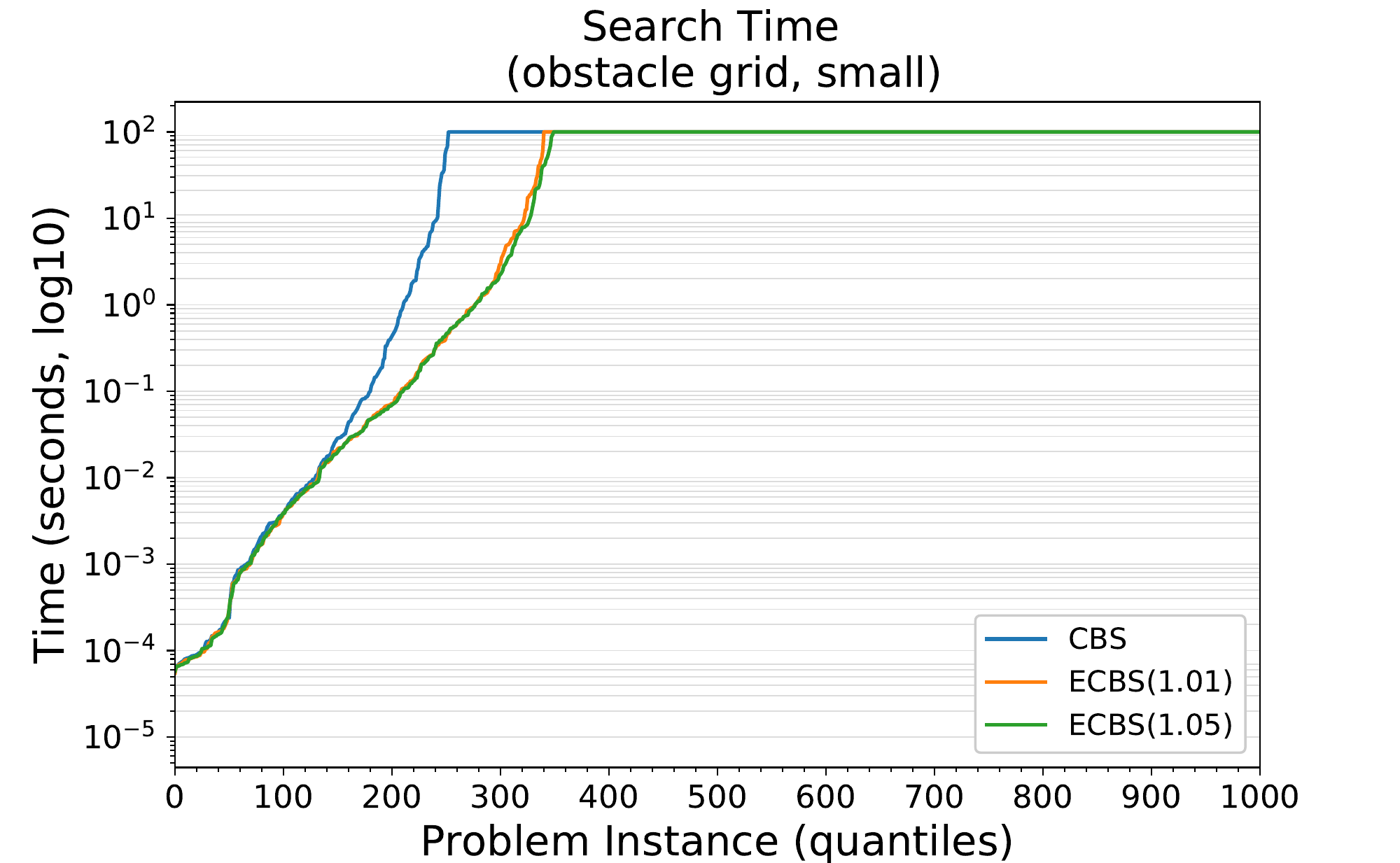}
    \begin{center}
    (e)
    \end{center}
    \end{minipage}
    \hspace*{-1.8em}
    \begin{minipage}{.37\textwidth}
    \includegraphics[width=\textwidth]{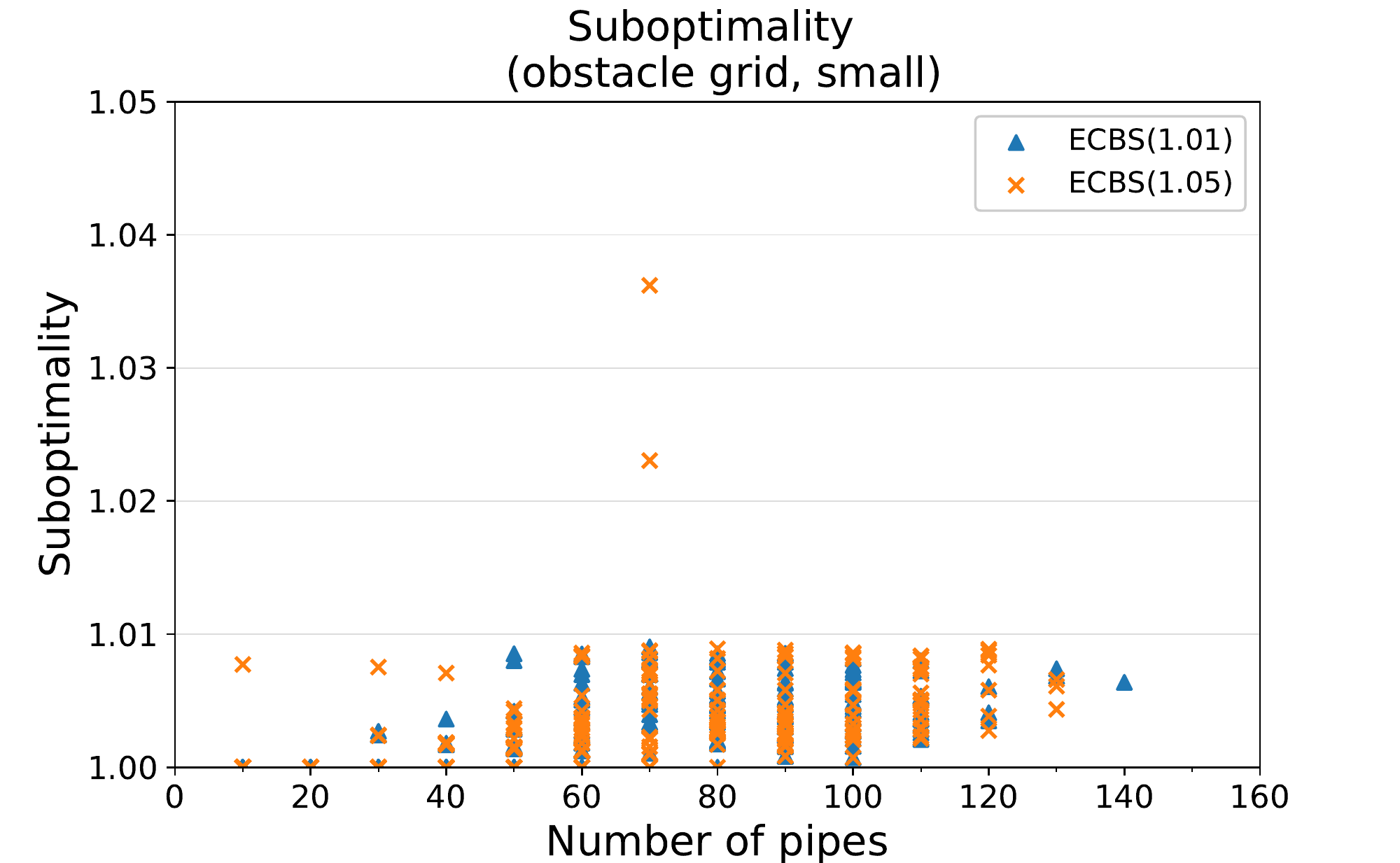}
    \begin{center}
    (f)
    \end{center}
    \end{minipage}
\end{minipage}
\label{fig:results_small}
\caption{Results on small environments of size $20 \times 20 \times 20$. Figures (a) - (c) show success rate, solution quality and runtime, respectively, for the empty variant, while
Figures (d) - (e) show the same metrics for the obstacles variant.
}\label{fig:sm}
\end{figure*}

In this experiment we focus on optimal and bounded-suboptimal pipe routing in a small congested environment of size $20 \times 20 \times 20$,
both with and without obstacles.
Figure~\ref{fig:sm} gives a summary. We see that: (i) CBS usually succeeds for problems of between 50 - 80 pipes 
but it times out on approximately 50\% of problem instances. Meanwhile ECBS usually succeeds for problems of between 100 - 130 pipes
and it manages to solve more than 75\% of all problem instances. 
Because the environment is small, there is little difference between the two bounded suboptimal solvers: there exist few solutions in the
FOCAL list of either method which are more than 1\% from optimal. This is also reflected in the results for solution quality, where
we observe few plans which are more than 1\% above the best known bound.

\begin{figure*}
\begin{minipage}{\textwidth}
    \hspace*{-1em}
    \begin{minipage}{.37\textwidth}
    \includegraphics[width=\textwidth]{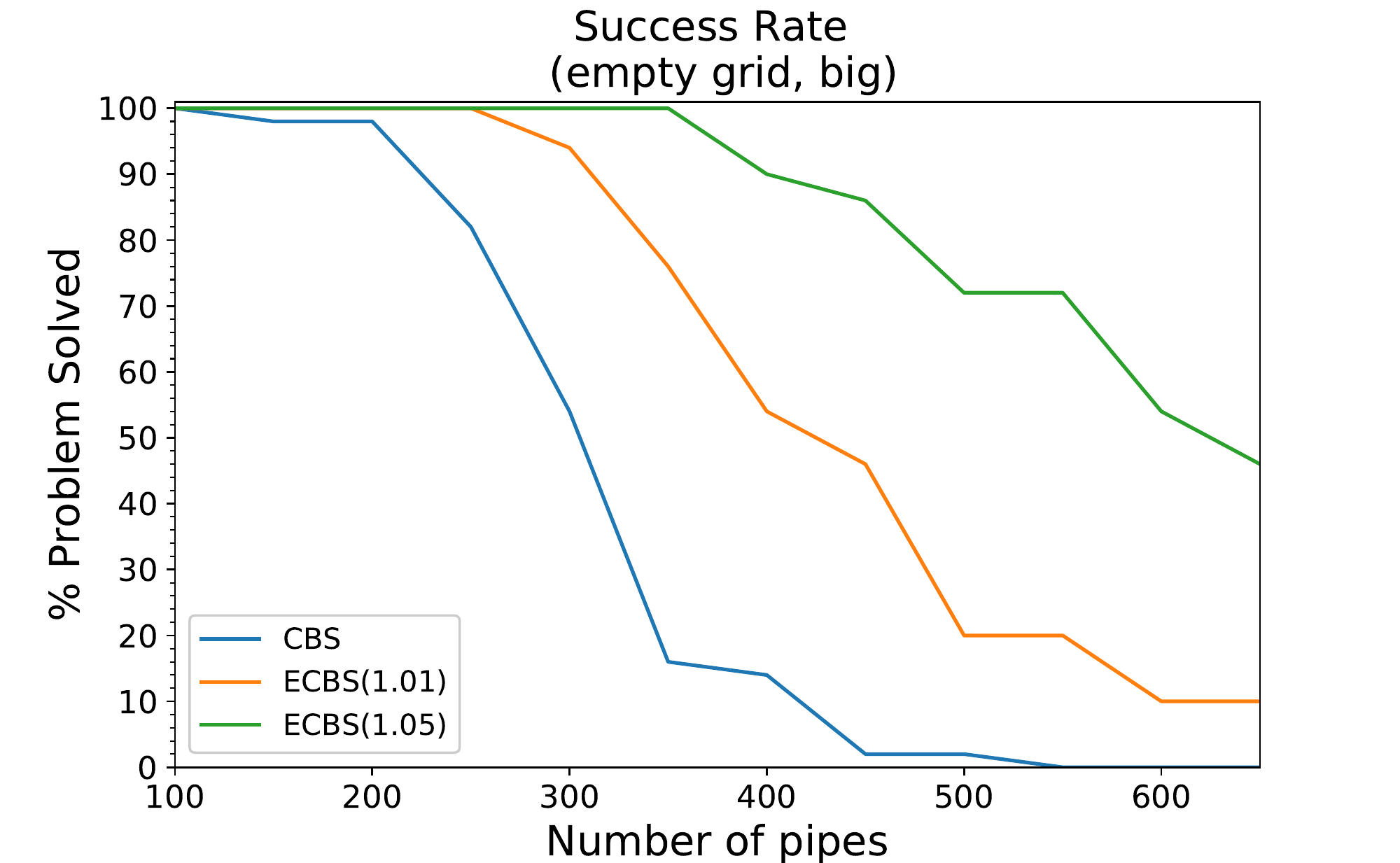}
    \begin{center}
    (a)
    \end{center}
    \end{minipage}
    \hspace*{-1.8em}
    \begin{minipage}{.37\textwidth}
    \includegraphics[width=\textwidth]{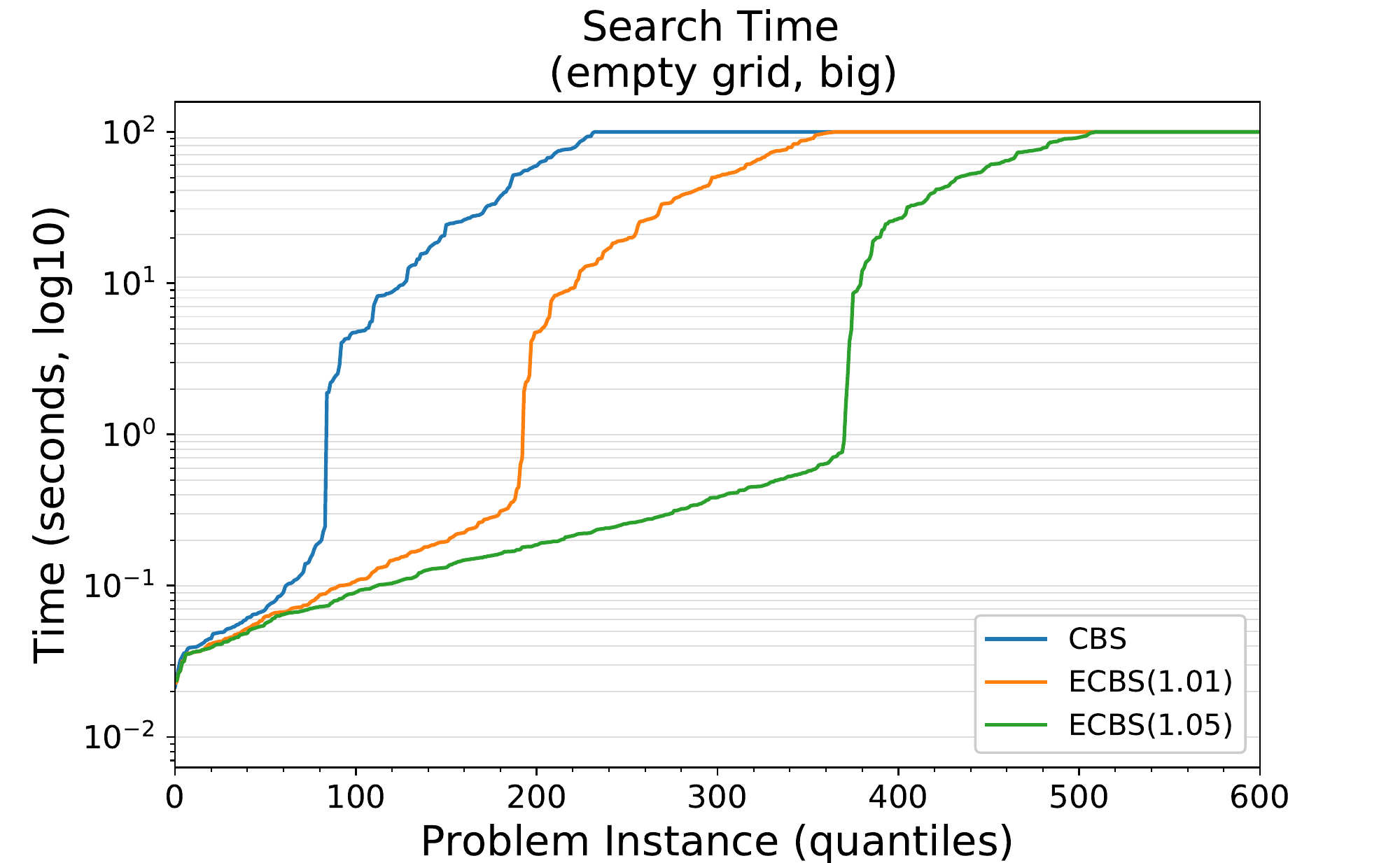}
    \begin{center}
    (b)
    \end{center}
    \end{minipage}
    \hspace*{-1.8em}
    \begin{minipage}{.37\textwidth}
    \includegraphics[width=\textwidth]{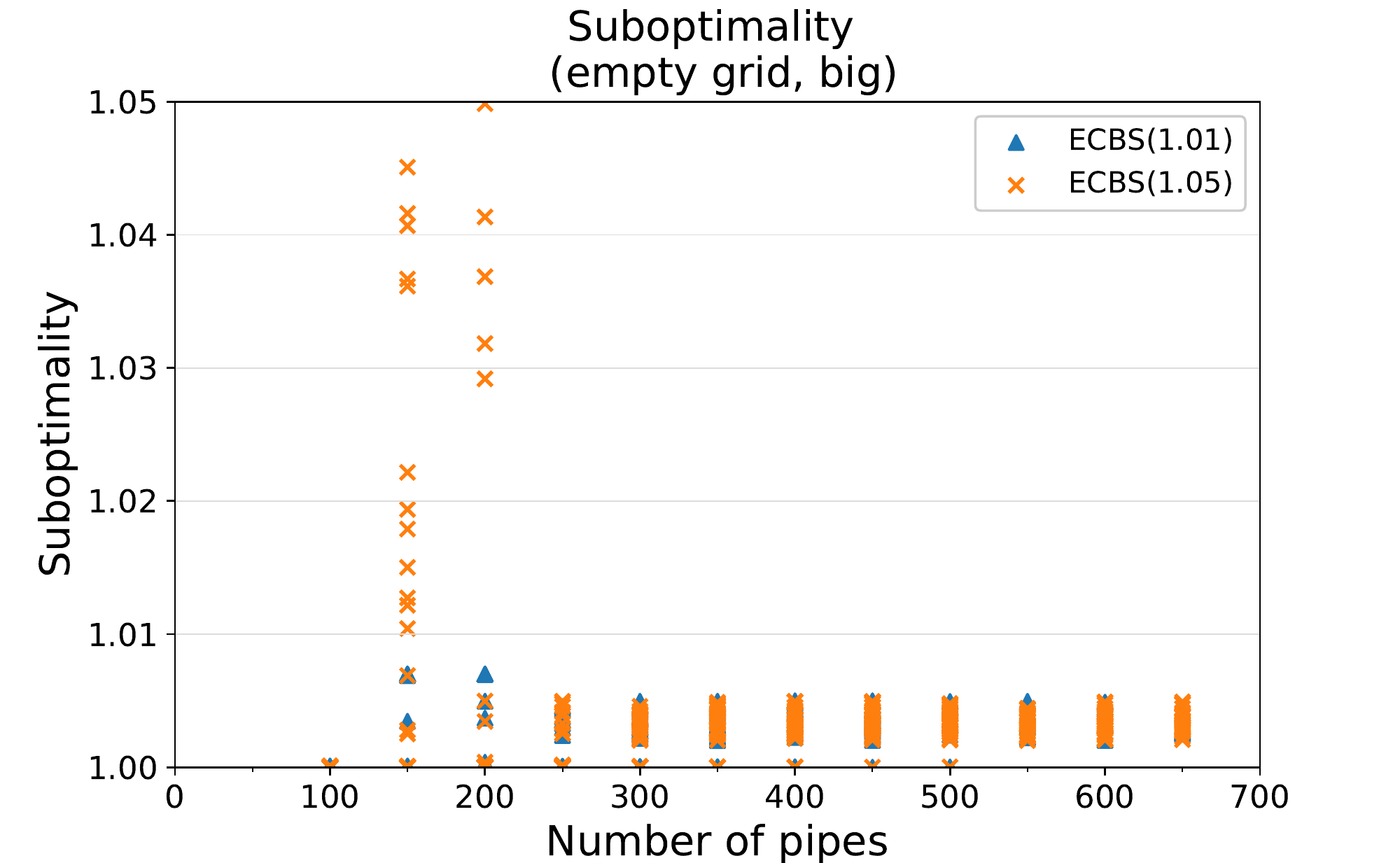}
    \begin{center}
    (c)
    \end{center}
    \end{minipage}
    \hspace*{-1em}
    \begin{minipage}{.37\textwidth}
    \includegraphics[width=\textwidth]{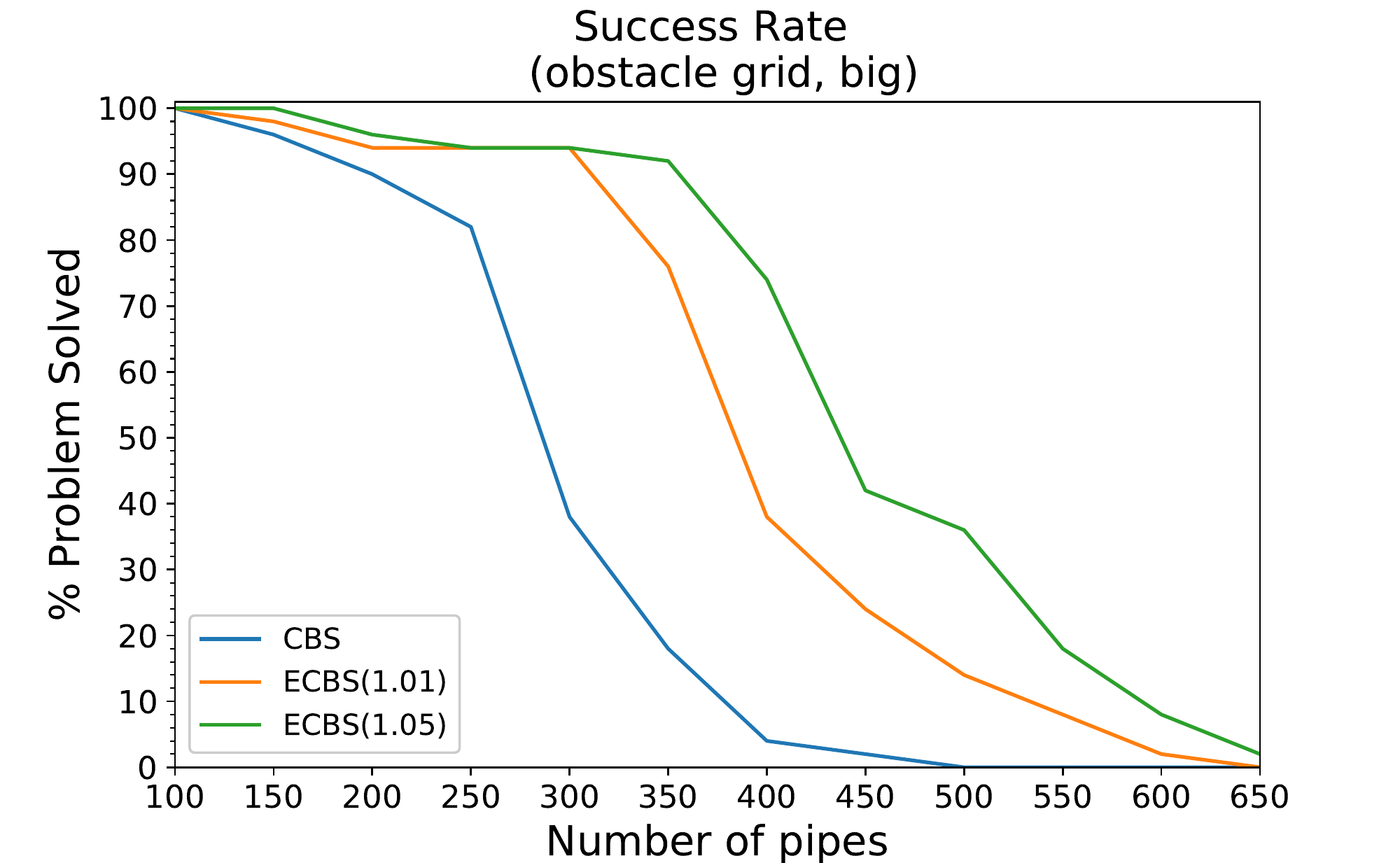}
    \begin{center}
    (d)
    \end{center}
    \end{minipage}
    \hspace*{-1.8em}
    \begin{minipage}{.37\textwidth}
    \includegraphics[width=\textwidth]{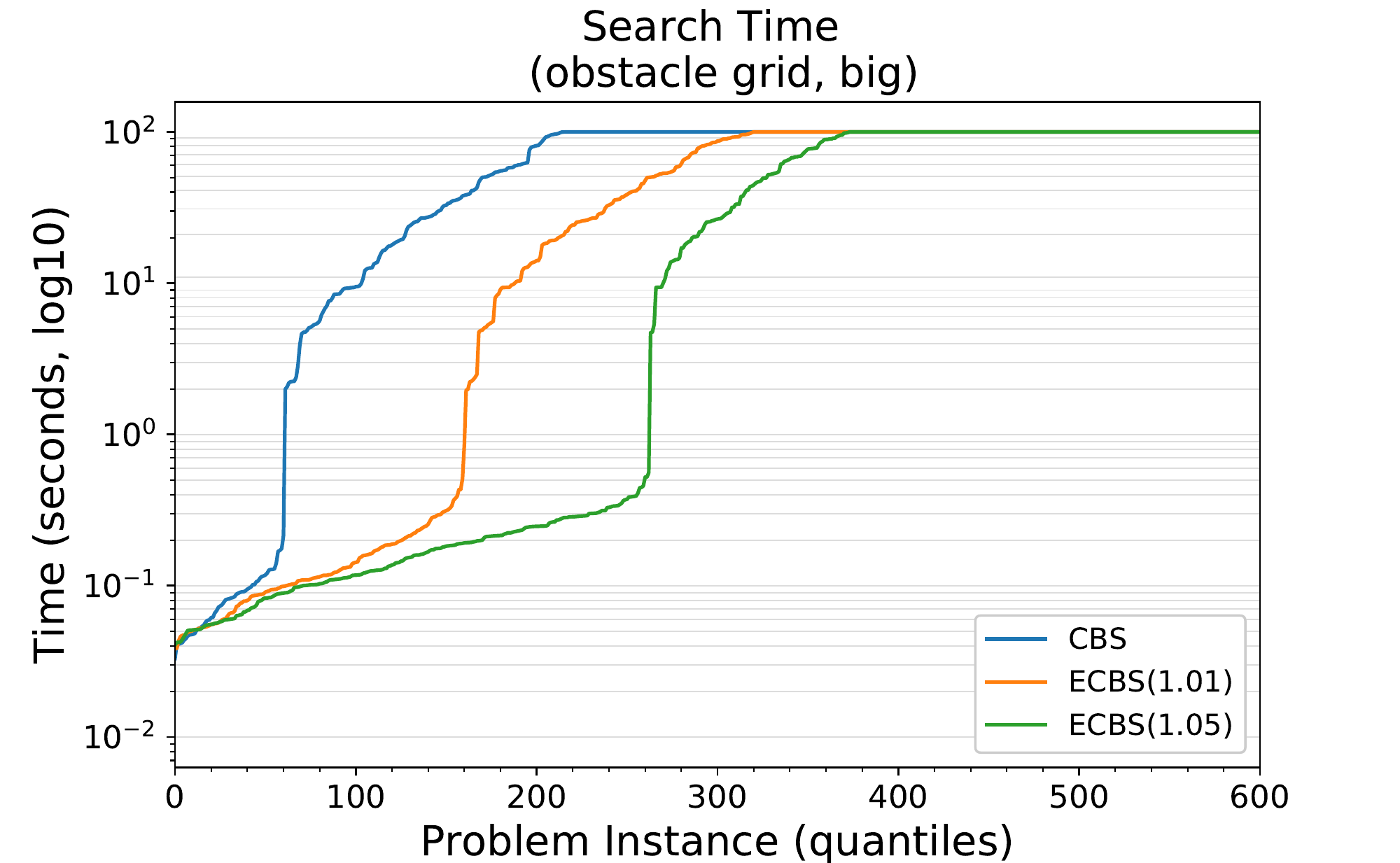}
    \begin{center}
    (e)
    \end{center}
    \end{minipage}
    \hspace*{-1.8em}
    \begin{minipage}{.37\textwidth}
    \includegraphics[width=\textwidth]{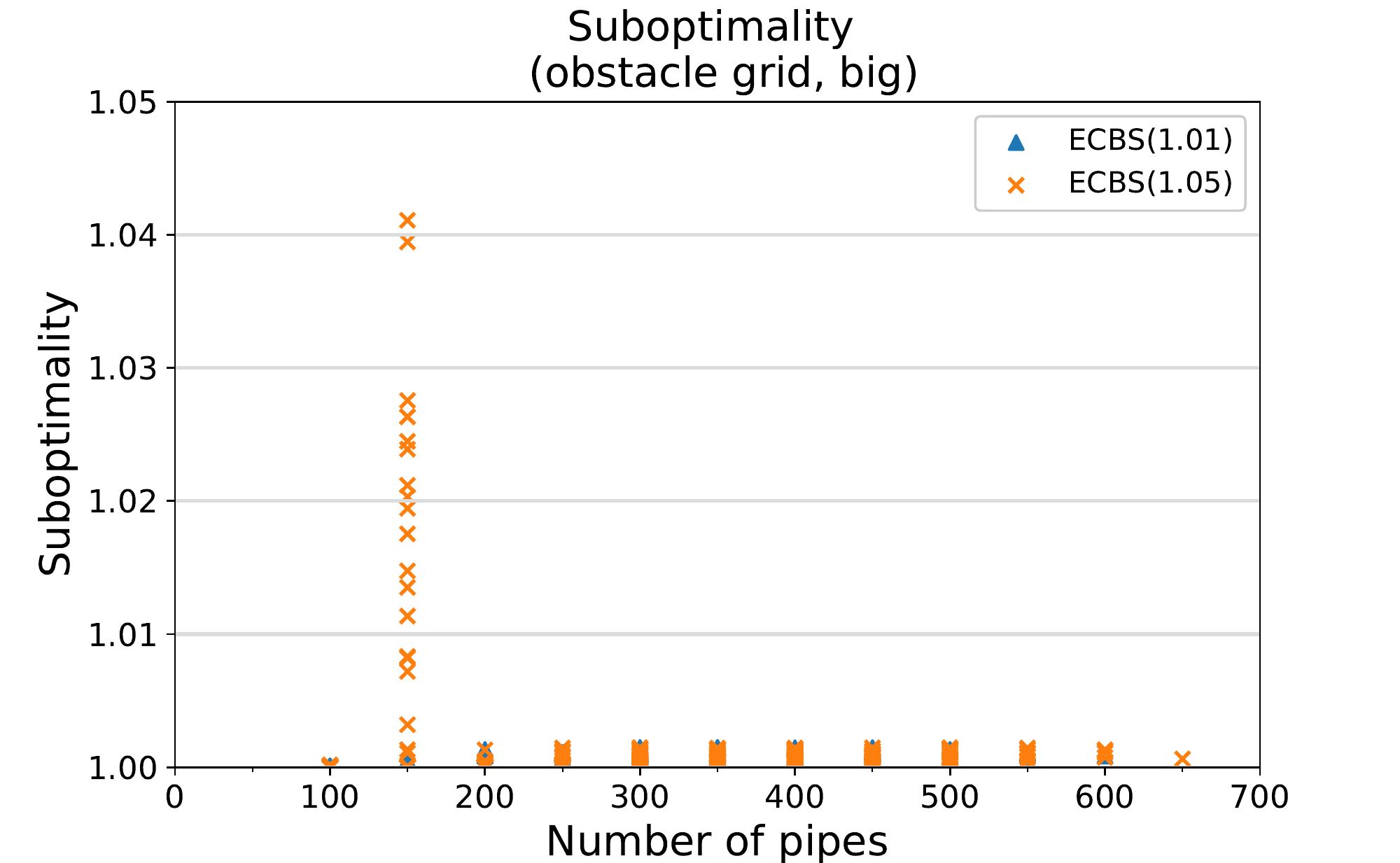}
    \begin{center}
    (f)
    \end{center}
    \end{minipage}
\end{minipage}
\label{fig:results_large}
\caption{Results on large environments of size $320 \times 320 \times 320$. Figures (a) - (c) show success rate, solution quality and runtime, respectively,  for the empty variant, while
Figures (d) - (e) show the same metrics for the obstacles variant.
}
\label{fig:lg}
\end{figure*}

\subsection{Experiment 2: Large Environments}
In this experiment we focus on pipe routing in a large environment of size 320 $\times$ 320 $\times$ 320. 
Such environments (with millions of cells) arise in settings that require pipes to be placed with a high-degree of 
precision (e.g. centimetres or less).
We again consider two variants, with and without obstacles.
Figure~\ref{fig:lg} gives a summary. 

Here we observe that even optimal CBS can usually solve instances with more than 300 pipes.
Interestingly, there is now a significant difference in runtime performance between ECBS(1.01) and ECBS(1.05),
with the latter usually solving problems that approach 450 pipes.
Both suboptimal variants find solutions that are within 1\% of the best known lower-bound, with
a small number of notable exceptions for ECBS(1.05) at the 150 and 200 pipe increments.

We observe also that runtime performance experiences a sharp phase transition. 
For example, between the 260th and 360th quantile of the obstacle environment, 
ECBS(1.05) runtime rises from a less than a second per instance to 
$\ge 100$ (i.e. timeout). In other words, the algorithm quickly finds feasible solutions, when
such exist, within the initial f-value suboptimality bound, but fails when more complex 
coordination is required. A higher initial bound may help ECBS to find feasible solutions 
quickly for the timeout cases. However, more work is needed to establish whether the cost of such
plans will remain significantly smaller than the promised bound.



%

%% file: discussion.tex
\section{The Real-World PR Problem}
\label{sec:pipe_routing}

As mentioned before, this paper focuses on the second phase of the process plant layout problem, where the equipment has already been positioned safely and correctly within the plant's volume, and the aim now is to determine the best  routing for the pipes that connect the equipment.

Focusing on the principles of MAPF algorithms, in this work we have ignored most of the real-world constraints and objectives, 
only requiring the pipes to traverse  the given feasible areas while minimizing their total length.
For completeness, and to motivate further research, below we describe the most important features of the real-world PR problem and the simplifications made by the PlantLayout prototype software of \citep{BelovEtAl2017}.


\noindent
\textbf{The Routing Landscape.} Obstacles can have any geometric shape. PlantLayout approximates them by cuboids.
\noindent
\textbf{Circular Bends and Non-Axis-Parallel Segments.} Pipes can bend along some minimal radius. Moreover, although in practice most pipe segments are parallel to one of the coordinate axes, a few are not. PlantLayout assumes pointed bends and axis-parallel segments.

\noindent
\textbf{Pipe Diameter.} Pipes have a certain diameter and require some minimal distance to other objects. PlantLayout accurately models this.

\noindent
\textbf{Support Costs.} A pipe has to be supported either by the ground or by some other equipment. The further it is from the supporting object or the ground, the higher the support costs. Moreover, in practice several pipes can be supported together when their routes are close. PlantLayout ignores the cost aspect by demanding that every pipe has to be at most 3 meters away from some component or the ground.

\noindent
\textbf{Stress and Flexibility Analysis.} Pipes may contract and expand due to temperature changes in the environment and the materials they transport. This poses stress on the pipe, which needs to be accounted for using stress and flexibility analysis. There are several methods differing in their complexity and precision \citep{ASMEB313}.


%% file: relatedwork.tex
\subsection{Existing Approaches for Real-World Pipe Routing}
\label{sec:relatedwork}



For small instances of plant layout design problem, \citet{Sakti2016} successfully apply an integrated approach for a \emph{satisfaction} version of the problem that \emph{simultaneously} allocates equipment and routes the pipes. In particular, they considered 10 equipment pieces and up to 15 pipes with 4 segments on average. However, they failed to find solutions for instances with as few as 8 pipes. Realistic plants are much larger:~\cite{BelovEtAl2017} consider one of five modules in a real liquid natural gas plant (the acid gas removal module), which already has 17 equipment pieces and 27 pipes within a 381;201;216 grid graph along axes $x;y;z$. 
For this and other larger, realistic problem instances that consider all equipment in the plant (containing often in excess of 100 equipment pieces and 200 pipes), 
integrated approaches do not scale, and the problem is divided into the two phases mentioned before: allocation and routing. 

Most research into plant layout design focuses only on the equipment allocation phase (e.g.,~\citealt{XuPapa07,XuPapa09}).
While some research includes the pipe routing phase (e.g.,~\citealt{Guirardello2005,Sakti2016}) or even focuses on it (e.g.,~\citealt{ZhuLatombe1991} and~\citealt{Jiang201563}), existing approaches do not consider simultaneous optimal routing.

\begin{figure}
    \centering
    \unitlength1pt
    \begin{picture}(100,50)
    \put(0,0){\vector(1,0){90}}
    \put(90,0){\makebox(0,0)[l]{ $X$}}
    \put(0,0){\vector(0,1){45}}
    \put(0,45){\makebox(0,0)[r]{$Z$ }}
    \multiput(5,20)(0,5){3}{
      \multiput(0,0)(5,0){2}{\makebox(5,5){$\times$}}
      \multiput(15,0)(5,0){5}{\makebox(5,5){$\times$}}
      \multiput(45,0)(5,0){3}{\makebox(5,5){$\times$}}
      \multiput(65,0)(5,0){2}{\makebox(5,5){$\times$}}
      \multiput(80,0)(5,0){2}{\makebox(5,5){$\times$}}
    }
    \multiput(5,10)(0,5){2}{
      \multiput(0,0)(5,0){18}{\makebox(5,5){$\times$}}
    }
    \put(45,22.5) 
    { 
      \thicklines
      \color{red}
      \put(0,0){\line(0,1){15}}
      \put(0,15){\line(1,0){35}}
      \put(35,15){\line(0,-1){15}}
    }
    \put(17.5,22.5){
      \put(0,0){\line(0,1){15}}
      \put(0,15){\line(1,0){50}}
      \put(50,15){\line(0,-1){15}}
    }
    \put(47.5,42.5){\makebox(3,5){$P_1$}}
    \put(67.5,42.5){\makebox(3,5){$P_2$}}
    \end{picture}
    \hfill
    \begin{picture}(100,50)
    \put(0,0){\vector(1,0){90}}
    \put(90,0){\makebox(0,0)[l]{ $X$}}
    \put(0,0){\vector(0,1){45}}
    \put(0,45){\makebox(0,0)[r]{$Y$ }}
    \put(45,22.5) 
    { 
      \thicklines
      \color{red}
      \put(0,0){\line(0,1){10}}
      \put(0,10){\line(1,0){35}}
      \put(35,10){\line(0,-1){10}}
    }
    \put(17.5,22.5){
      \put(0,0){\line(0,-1){10}}
      \put(0,-10){\line(1,0){50}}
      \put(50,-10){\line(0,1){10}}
    }
    \put(47.5,22.5){\makebox(0,0)[r]{$P_1$}}
    \put(67.5,22.5){\makebox(3,5)[b]{$P_2$}}
    \end{picture}
    \caption{The left picture shows a layout in $XZ$-projection (side-view);
    the right one in $XY$ (top-view). Blocked cells are marked with $\times$
    on the left; assume that this obstacle structure is replicated along the
    $Y$ direction. There are two pipes, red and black, with their start-goal
    locations in holes 1-3  and 2-4, respectively. $P_1$ and $P_2$ mark
    conflict points. Fixed priority planning always fails in this case,
    irrespective of the ordering. The optimal solution 
    (right picture) has the two pipes swap priorities at the conflict points.
    }
    \label{fig:PBS}
\end{figure}
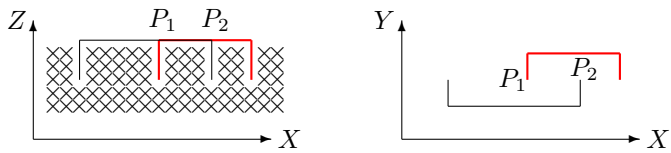

On the one hand, the more realistic approaches, which do take into account the simultaneous routing of several pipes (including branching pipes and support placement), are based on heuristic algorithms (rather than complete search methods), such as the ant-colony evolutionary algorithms used by~\cite{Furuholmen2010,Jiang201563}. 
One popular approach involves prioritised planning~\citep{cao2018design} wherein the pipes are ordered according to some fixed priority and routed in order. Each higher priority pipe then becomes an obstacle for all lower-priority pipes. The main disadvantages of prioritised planning are suboptimality and incompleteness in general. To illustrate the limitations consider the example in Figure~\ref{fig:PBS}.

On the other hand, the approaches that rely on complete search methods are either difficult to extend for some of the required constraints, or quickly become intractable. 
One of the most realistic of the complete approaches is that of \cite{Guirardello2005}, which provides a detailed mixed integer programming (MIP) model for solving phase one, and a general overview of a network-flow MIP model for solving phase two. This second MIP model relies on the construction of a reduced connection graph that limits the possible routes of the pipes. This is used to route pipes one by one, since they suggest that simultaneous routing is too costly. 
While they do not give enough details regarding how the connection graph is constructed, an approach to construct such a connection graph is given, e.g., by \cite{SPP_ND_1992}, who present a higher-dimensional rectilinear shortest path model that considers bend costs. A more hierarchical method using cuboid free space decomposition is given by \cite{ZhuLatombe1991} and applied to pipe routing. However, even if these methods are used, it is not clear how~\cite{Guirardello2005} perform sequential pipe routing when pipes interfere with each other (\citealt{Guirardello2005} talk about ``some tuning by hand'' which might be required for these cases). 

\cite{BelovEtAl2017} describe a general pipe routing method that incorporates stress analysis. While their model can be applied to route several pipes simultaneously, it quickly becomes intractable. In contrast, as shown in our experiments, recent advances in MAPF allow us to solve problems with hundreds of pipes optimally or very nearly optimal (and always with a strong cost guarantee).

%% file: conclusion.tex
\section{Conclusion}
\label{sec:conclusion}
In this paper we have shown that the 3D  Pipe  Routing  (PR)  problem,  which  aims  at  placing
collision-free pipes from given start locations to given goal
locations  in  a  known  3D  environment, is similar to the 2D MAPF problem. This is important because it indicates  that many recently developed MAPF algorithms apply more broadly than currently believed in the MAPF research community. To demonstrate this, we have evaluated the success rate, solution quality and efficiency of three MAPF algorithms (CBS, ECBS(1.01) and ECBS(1.05)) in several different environments. Results show that MAPF algorithms are able to find solutions for large instances with optimal or near optimal quality. This provides strong incentives to the MAPF research community to perform the further  research necessary to tackle real-world pipe-routing instances of interest to industry today.